\documentclass[10pt]{article}
\usepackage[margin=1in]{geometry}
\usepackage{amsmath,amssymb,amsfonts}
\usepackage{booktabs}
\usepackage{graphicx}
\usepackage{xcolor}
\usepackage[numbers,sort&compress]{natbib}
\usepackage[colorlinks=true,linkcolor=blue!60!black,citecolor=blue!60!black,urlcolor=blue!60!black]{hyperref}
\usepackage{tikz}
\usetikzlibrary{positioning,arrows.meta,fit,backgrounds}
\usepackage{caption}
\usepackage{microtype}
\usepackage{placeins}

\newcommand{\holdMAE}{4.90}
\newcommand{\oursMAE}{3.24}
\newcommand{\oursOneLoopMAE}{3.24}
\newcommand{\oursSteps}{41{,}790}
\newcommand{\oursLatThree}{94}
\newcommand{\oursLatOne}{84}

\newcommand{\oursEncOne}{14}
\newcommand{\oursEncThree}{30}
\newcommand{\oursBestMAE}{3.22}
\newcommand{\oursBestSteps}{35{,}000}
\newcommand{\ablMAE}{3.59}
\newcommand{\ablSteps}{76{,}893}
\newcommand{\ablBestMAE}{3.26}
\newcommand{\ablBestSteps}{25{,}000}
\newcommand{\actMAE}{3.22}
\newcommand{\actSteps}{100{,}000}
\newcommand{\actLat}{20}
\newcommand{\dpMAE}{3.21}
\newcommand{\dpSteps}{75{,}580}
\newcommand{\dpLat}{960}
\newcommand{\smolMAE}{3.03}
\newcommand{\smolSteps}{27{,}638}

\newcommand{\instrAMAE}{3.84}
\newcommand{\instrAN}{256}
\newcommand{\instrBMAE}{2.32}
\newcommand{\instrBN}{148}
\newcommand{\instrCMAE}{3.23}
\newcommand{\instrCN}{1596}
\newcommand{\ablPeakVal}{3.26}
\newcommand{\oursPeakVal}{3.18}
\newcommand{\ablEndVal}{3.59}

\newcommand{\replayMAE}{3.76}
\newcommand{\replayHold}{4.78}
\newcommand{\replayN}{6}
\newcommand{\replayTypEp}{520}
\newcommand{\replayTypMAE}{2.76}
\newcommand{\replayWorstEp}{318}
\newcommand{\replayWorstMAE}{8.49}

\title{VeloLoop: Mixture-of-Loops Observation Encoders for\\Flow-Matching Manipulation Policies}
\author{A Adith \\ \texttt{aditharavind03@gmail.com}}
\date{}

\begin{document}
\maketitle

\begin{abstract}
We present \textbf{VeloLoop}, a vision-language-action (VLA) policy whose observation encoder is a \emph{looped}
Transformer built on the Mixture-of-Loops Transformer (MoLT)~\cite{zhu2026molt}. One weight-shared stack is applied
several times to the camera, language and proprioceptive tokens. Every loop restarts from the input embeddings and
receives feedback only as key-value memory from the previous loop, read under a single softmax with the loop's own
per-layer key-value pairs. A cross-only decoder serves as a flow-matching action expert and never feeds back into the
loop. We adapt MoLT's strict-prefix memory rule to bidirectional observation contexts and train with \emph{loop
forcing}, an action loss after every loop. On 1{,}116 teleoperated pick-and-place demonstrations on a 6-DoF tabletop manipulator, under one
shared protocol, VeloLoop reaches an arm error of \oursMAE$^\circ$ over 1.6\,s action chunks on 56 held-out episodes
(holding the current pose: \holdMAE$^\circ$), on par with ACT (\actMAE$^\circ$) and Diffusion Policy
(\dpMAE$^\circ$) and behind the robot-pretrained, 450M-parameter SmolVLA (\smolMAE$^\circ$). Looping acts as a
training-time regulariser: a parameter-matched non-looped ablation overfits, from \ablBestMAE$^\circ$ to
\ablMAE$^\circ$, while VeloLoop stays within $0.05^\circ$ of its best, and it is lower at every equal step count
from 20k steps on. Extra loops do not help at inference; thanks to loop forcing, one loop matches three, so the
benefit comes at the inference cost of a plain encoder. Closed-loop on the real arm, VeloLoop ran an uninterrupted
11-minute session, holding the block in 17 of 25 grasp attempts and placing it at the box 14 times, at 35\,ms per
48-step chunk.
\end{abstract}

\section{Introduction}
VLA policies such as RT-2~\cite{brohan2023rt2}, OpenVLA~\cite{kim2024openvla}, $\pi_0$~\cite{black2025pi0} and
SmolVLA~\cite{shukor2025smolvla} couple a pretrained vision-language backbone with an action head and grow mainly by
adding parameters. Robot deployment, however, is tightly constrained in memory and latency: the policy shares one
on-board GPU with perception and control and must answer at 10--30\,Hz. Looped (weight-tied) Transformers decouple
depth from parameter count by applying the same layers repeatedly
\cite{dehghani2019universal,giannou2023looped,geiping2025scaling}, and several 2026 works bring looping to robot
policies~\cite{tur2026rdvla,shen2026loopvla,dao2026loopedwam}. All of them carry a hidden state between iterations.

The Mixture-of-Loops Transformer (MoLT)~\cite{zhu2026molt} passes something else between loops. In its loop-KV
form, every loop restarts from the input embeddings and receives feedback only as key-value memory projected from
the previous loop's output, read together with the current loop's per-layer KV under one softmax (Mixture-of-Loops
Attention, MoLA). Its U-shaped variant separates memory construction, by a looped encoder, from readout, by a
cross-only decoder. This split matches a flow-matching VLA: the observation context is a fixed token set to be
understood once per control step, and the action expert reads it at every denoising step. We therefore build
\textbf{VeloLoop}, a looped MoLT encoder over the observation context with a non-looped MoLT decoder as the
flow-matching action expert, and ask what looping buys on a real-robot dataset. Our contributions are:
\begin{itemize}
  \item To our knowledge, the first robot policy built on MoLT's loop-KV paradigm, with MoLA adapted from causal
        language modelling to unordered multimodal contexts (memory read from every \emph{other} position, $j\neq t$)
        and all looping placed in the context encoder, so loops do not multiply with denoising steps.
  \item Evidence that looped training with loop forcing is a regulariser: against a parameter-matched non-looped
        ablation it removes overfitting, while extra loops add no accuracy at inference. Because every loop is
        supervised, the same checkpoint can be deployed with a single loop at the cost of a plain encoder.
  \item A controlled comparison with ACT~\cite{zhao2023act}, Diffusion Policy~\cite{chi2023diffusionpolicy} and
        SmolVLA~\cite{shukor2025smolvla} on 1{,}116 demonstrations from a 6-DoF tabletop manipulator (one data pipeline, split, chunk length
        and held-out evaluation), and an 11-minute closed-loop session on the arm with every grasp attempt reported.
\end{itemize}

\section{Related Work}
\label{sec:related}
\paragraph{VLA policies and action heads.}
RT-1~\cite{brohan2023rt1} and RT-2~\cite{brohan2023rt2} established large Transformer policies and VLM co-fine-tuning;
Open X-Embodiment~\cite{oneill2024oxe} and Octo~\cite{ghosh2024octo} scaled cross-embodiment data, and
OpenVLA~\cite{kim2024openvla} and OpenVLA-OFT~\cite{kim2025oft} made open 7B VLAs fast. $\pi_0$~\cite{black2025pi0}
attached a flow-matching expert to a PaliGemma VLM~\cite{beyer2024paligemma}, the template our decoder follows,
extended by $\pi_{0.5}$~\cite{intelligence2025pi05} and FAST~\cite{pertsch2025fast}; diffusion and flow heads also
appear in CogACT~\cite{li2024cogact}, RDT-1B~\cite{liu2025rdt}, GR00T N1~\cite{bjorck2025gr00t} and
FLOWER~\cite{reuss2025flower}. SmolVLA~\cite{shukor2025smolvla} is the closest small-scale VLA. ACT~\cite{zhao2023act}
predicts action chunks with a CVAE, Diffusion Policy~\cite{chi2023diffusionpolicy} denoises them, and flow
matching~\cite{lipman2023flow,liu2023rectified} allows few-step sampling, which we use.

\paragraph{Looped Transformers.}
Weight sharing across depth goes back to the Universal Transformer~\cite{dehghani2019universal}, adaptive
halting~\cite{graves2016adaptive} and ALBERT~\cite{lan2020albert}. Looped Transformers can emulate
programs~\cite{giannou2023looped}, learn in-context algorithms~\cite{yang2024looped} and simulate latent
reasoning~\cite{saunshi2025reasoning}; recurrent-depth language models~\cite{geiping2025scaling,
mcleish2025retrofitted,zhu2025ouro,bae2025relaxed,bae2025mixture,csordas2024moeut} scale test-time compute, mostly by
carrying the residual stream between loops. Attention across depth~\cite{pagliardini2024denseformer,zhu2026moda} and
self-/cross-decoder KV reuse~\cite{sun2024yoco,sun2026universalyoco} lead to MoLT~\cite{zhu2026molt}. LoopFormer
\cite{jeddi2026loopformer}, ELT~\cite{goyal2026elt} and LiFT~\cite{derakhshani2026lift} train models usable at many
loop counts.

\paragraph{Looped and adaptive-compute robot policies.}
RD-VLA~\cite{tur2026rdvla} loops a recurrent action head with random-depth training; LoopVLA~\cite{shen2026loopvla}
loops a few shared VLM layers with per-iteration supervision and a halting score, in simulation; concurrently,
LoopedWAM~\cite{dao2026loopedwam} loops both experts of a world-action model on a real 6-DoF arm. Smaller recurrent
policies include HR-AH~\cite{makineni2026hrah}, Tiny Recursive Policy~\cite{aktas2026trp}, MPCoT~\cite{zhang2026mpcot}
and ARVLAT~\cite{allam2026arvlat}; in RL, looped policies benefit from extra compute~\cite{ghugare2026computation,
rusch2026loopedactor} but specialise to their training depth unless trained to be elastic~\cite{wang2026flexloop}.
Non-looped VLAs adapt compute by early exit or layer skipping (DeeR-VLA~\cite{yue2024deer} on
RoboFlamingo~\cite{li2024roboflamingo}, MoLe-VLA~\cite{zhang2026mole}, DySL-VLA~\cite{yang2026dysl},
A1~\cite{zhang2026a1}, AC$^2$-VLA~\cite{yu2026ac2vla}, Think-Less-Act-Early~\cite{lei2026thinkless}), denoising
budgets~\cite{li2026elastic,wang2026eqrl,salla2026pogp}, caching~\cite{xu2025vlacache}, pruning~\cite{yang2025efficientvla}
or asynchronous execution~\cite{black2025rtc}. VeloLoop differs in looping the observation encoder with embedding
restart and KV memory instead of a carried state, and in keeping the action expert unlooped.

\section{Method}
\label{sec:method}

\begin{figure}[t]
\centering
\begin{tikzpicture}[
  box/.style={draw, rounded corners=2pt, minimum height=0.7cm, align=center, font=\footnotesize, fill=white},
  arr/.style={-{Latex[length=2mm]}, thick},
  darr/.style={-{Latex[length=2mm]}, thick, dashed}]
  \node[box, minimum width=1.9cm] (g) at (0, 1.35) {global camera};
  \node[box, minimum width=1.9cm] (w) at (0, 0.45) {wrist camera};
  \node[box, minimum width=1.9cm] (t) at (0,-0.45) {instruction};
  \node[box, minimum width=1.9cm] (s) at (0,-1.35) {joint state};
  \node[box, text width=2.1cm, minimum height=3.3cm] (enc) at (2.9,0) {SigLIP vision\\(+ $2{\times}2$ merge)\\[2pt]SigLIP text\\(frozen)\\[2pt]state MLP};
  \node[box, text width=1.5cm, minimum height=1.6cm] (E) at (5.25,0) {context $E$\\111 tokens};
  \node[box, fill=blue!8, minimum width=1.4cm] (l1) at (7.6, 1.3) {loop 1};
  \node[box, fill=blue!8, minimum width=1.4cm] (l2) at (7.6, 0) {loop 2};
  \node[box, fill=blue!8, minimum width=1.4cm] (l3) at (7.6,-1.3) {loop 3};
  \node[box, text width=2.3cm, minimum height=3.3cm, fill=orange!10] (dec) at (12.4,0) {MoLT decoder\\(action expert)\\[2pt]flow matching,\\10 Euler steps\\[2pt]$48\times7$ chunk};
  \foreach \n in {g,w,t,s} \draw[arr] (\n.east) -- (\n.east -| enc.west);
  \draw[arr] (enc) -- (E);
  \foreach \n in {l1,l2,l3} \draw[arr] (E.east) -- (\n.west);
  \draw[darr] (l1.south) -- node[right, font=\scriptsize, xshift=1pt]{mem.\ KV} (l2.north);
  \draw[darr] (l2.south) -- node[right, font=\scriptsize, xshift=1pt]{mem.\ KV} (l3.north);
  \draw[arr, gray] (l1.east) -- node[above, font=\scriptsize, pos=0.55]{$z^1$} (l1.east -| dec.west);
  \draw[arr, gray] (l2.east) -- node[above, font=\scriptsize, pos=0.55]{$z^2$} (l2.east -| dec.west);
  \draw[arr] (l3.east) -- node[above, font=\scriptsize, pos=0.55]{$z^3$} (l3.east -| dec.west);
  \begin{scope}[on background layer]
    \node[draw, dashed, rounded corners, fit=(l1)(l3), inner sep=4pt, label={[font=\scriptsize]above:shared MoLA stack}] {};
  \end{scope}
\end{tikzpicture}
\caption{VeloLoop. One weight-shared MoLA stack is applied for $N$ loops to the observation context. Each loop
restarts from the embeddings $E$ and reads memory KV projected from the previous loop's normalised output. The
action expert, a cross-only MoLT decoder, reads decoder KV projected from a loop's output $z^r$. In training it is
supervised on every loop (loop forcing); at inference it reads only the last executed loop, so the same weights can
run with fewer loops.}
\label{fig:arch}
\end{figure}
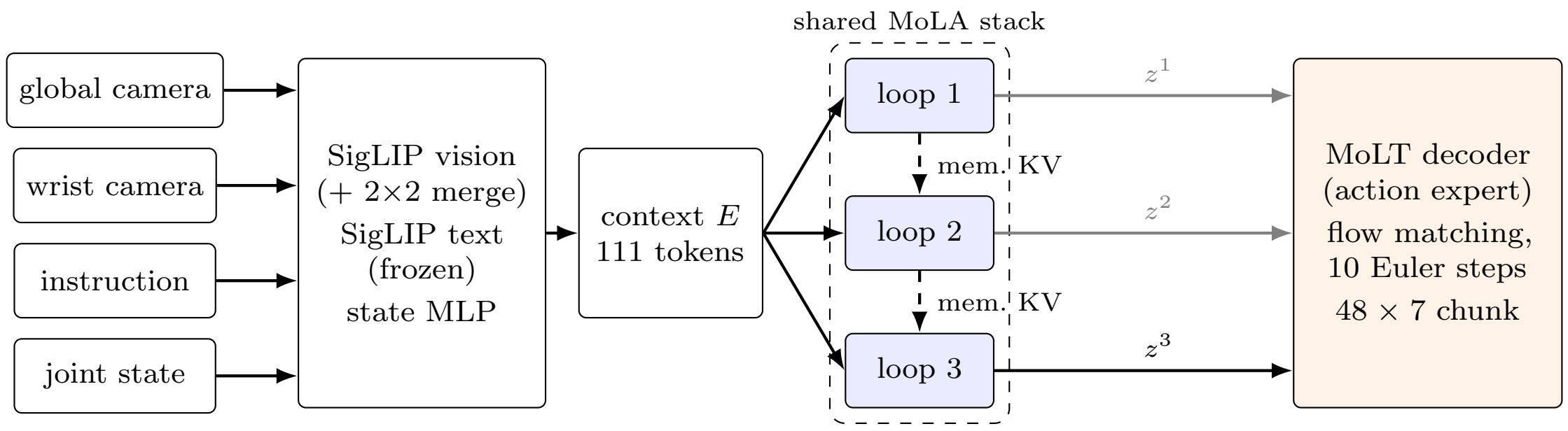

\paragraph{Problem setting.}
At each control step the policy observes a global and a wrist RGB image, a language instruction $\ell$ and the joint
state $q_t\in\mathbb{R}^7$, and predicts a chunk of $H=48$ joint targets $A_t=(a_t,\dots,a_{t+H-1})$ at 30\,Hz
(1.6\,s). We model $p(A_t\mid o_t)$ with conditional flow matching~\cite{lipman2023flow,liu2023rectified}: with
$x_\tau=\tau\epsilon+(1-\tau)A$ and $\epsilon\sim\mathcal{N}(0,I)$, a network $v_\theta$ regresses $\epsilon-A$.

\paragraph{Observation context.}
Both images are encoded by a SigLIP-B/16~\cite{zhai2023siglip} vision tower at $224\times224$ (first six layers
frozen), and a learned $2\times2$ patch-merge per camera reduces each $14\times14$ grid to 49 tokens. The frozen SigLIP
text tower encodes the instruction once (up to 12 tokens) and an MLP embeds the state as one token. With learned
position embeddings this gives the context $E\in\mathbb{R}^{111\times d}$, $d=768$.

\subsection{MoLT encoder over observation contexts}
The encoder is an $L_E=6$-layer pre-norm Transformer with SwiGLU blocks and per-head QK RMSNorm~\cite{zhu2026molt},
applied $N$ times with shared weights. Let $X^{r,\ell}$ be the residual stream entering layer $\ell$ in loop $r$ and
$z^r=\mathrm{Norm}(X^{r,L_E})$ the loop output. Every loop restarts from the embeddings, $X^{r,0}=E$, and each layer
applies Mixture-of-Loops Attention, one softmax over two source sets:
\begin{equation}
o_t^{r,\ell} = \sum_{(k,v)\in \mathcal{S}_t^{r,\ell}\,\sqcup\,\mathcal{D}_t^{r,\ell}}
\frac{\exp(q_t^\top k/\sqrt{d_h})}{\sum_{(k',v')\in\mathcal{S}_t^{r,\ell}\sqcup\mathcal{D}_t^{r,\ell}}\exp(q_t^\top k'/\sqrt{d_h})}\,v .
\end{equation}
\emph{Sequence sources} $\mathcal{S}$ are, for $r>1$, memory KV projected from the previous loop's output,
$k_j=\mathrm{Norm}(z^{r-1}_jW_{K,\mathrm{mem}}^\ell)$ and $v_j=z^{r-1}_jW_{V,\mathrm{mem}}^\ell$ (in loop 1, the layer's
own KV). \emph{Depth sources} $\mathcal{D}$ are the current loop's KV at the same token $t$ from all layers
$0\le k\le\ell$.

\paragraph{Bidirectional loop KV.} MoLT, a causal language model, reads memory only from a strict prefix, $j<t$,
which removes any dependence of token $t$ on its own previous-loop state. An observation context is unordered and
must be read bidirectionally. We keep the property that matters, no same-position cross-loop dependence, by reading
memory at every valid position except the query's own, $j\neq t$; token $t$'s own information then enters only
through depth sources built from the restarted embedding. As in the reference implementation, loop memory is
detached from the gradient. MoLA is computed exactly in a single fused attention call (Appendix~\ref{app:impl}).

\subsection{Action expert and loop forcing}
Following U-MoLT, the action expert is a cross-only decoder outside the feedback path. Its $H$ tokens embed the noisy
chunk $x_\tau$, the flow time $\tau$ and a horizon position, and each of its $L_D=6$ layers attends jointly to decoder
KV projected from $z^r$ and to the action tokens' own KV. The encoder therefore runs once per control step and only
the decoder is iterated over the 10 Euler steps. All loops share the same noise, flow time and targets, and the loss
supervises every loop:
\begin{equation}
\mathcal{L}=\sum_{r=1}^{N}\alpha_r\,\big\|\,v_\theta(x_\tau,\tau\mid z^r)-(\epsilon-A)\big\|_2^2,
\qquad \alpha=(1,2,3)/6 ,
\end{equation}
masked beyond the end of the episode. Loop forcing trains the feedback recurrence in parallel, as in MoLT, and asks
the shared stack to produce a usable representation after every pass, which we find acts as a regulariser
(Section~\ref{sec:results}). A side effect is that every prefix of the loop schedule is a valid policy, so a model
trained with $N$ loops can be deployed with fewer without retraining.

\section{Experimental Setup}
\label{sec:setup}
\paragraph{Data.} We use Tabletop-1.1k, 1{,}116 teleoperated demonstrations on a 6-DoF tabletop manipulator with a parallel gripper,
recorded at 30\,Hz with a global ($1280\times720$) and a wrist camera ($640\times480$). Episodes last 10--20\,s and
cover three instructions: ``pick the yellow block put into the box'' (854 episodes), ``pick the black block put into
the box'' (159) and ``pick the yellow block'' (100). We hold out 56 episodes (fixed random permutation, seed 0) and
remove three training episodes with recording glitches, leaving 1{,}057. Observations are sampled at 10\,Hz and
resized to $224\times224$ once into a shared cache used by every method.

\paragraph{Shared protocol.} All methods share the cache, split, quantile normalisation, GPU augmentation (random
shift, colour jitter), state noise and dropout, a single current observation and the 48-step chunk. ACT, Diffusion
Policy and SmolVLA use the LeRobot~\cite{cadene2026lerobot} implementations with default optimiser settings; SmolVLA
is fine-tuned from the public \texttt{smolvla\_base} checkpoint. The ablation is VeloLoop with $N=1$: the same
parameters and recipe, without looping. Each method trained within a fixed wall-clock budget on one shared GPU
(Table~\ref{tab:main}); the two VeloLoop runs (3 loops and the 1-loop ablation) had 10.9\,h each, with cosine schedules budgeted in wall-clock time.
Training details are in Appendix~\ref{app:impl}.

\paragraph{Checkpoints and metrics.} The main comparison uses, for every method, the checkpoint at the end of its
budget. VeloLoop and the ablation were also evaluated every 2{,}500 steps on 1{,}024 fixed held-out samples; the best
of these is reported separately as validation-selected. On 2{,}000 held-out samples (shared indices, fixed noise,
default sampling steps) we report the arm-joint MAE in degrees over the whole chunk and over steps 1--8 and 33--48,
the motion $R^2$ (share of the demonstrated displacement $a_{t+k}-q_t$ explained), gripper open/closed accuracy and
batch-1 latency. ``Hold current pose'' ($a_{t+k}=q_t$) is the error-scale reference.

\section{Results}
\label{sec:results}

\begin{table}[t]
\centering
\caption{Held-out open-loop accuracy on Tabletop-1.1k (56 episodes, 2{,}000 samples, 48-step chunks). MAE is in
degrees over the six arm joints; 1--8 and 33--48 are the MAE over those chunk steps, $R^2$ is the motion $R^2$ and
Grip.\,\% the gripper open/closed accuracy. Params in millions. Upper block: checkpoint at the end of each method's
budget. Last column: median batch-1 latency (ms) on a shared GPU, including all sampling steps.}
\label{tab:main}
\resizebox{\linewidth}{!}{\begin{tabular}{lrrrrrrrrr}
\toprule
Method & Params & Steps$\times$batch & GPU-h & MAE$\downarrow$ & 1--8$\downarrow$ & 33--48$\downarrow$ & $R^2\uparrow$ & Grip.\,\%$\uparrow$ & ms \\
\midrule
Hold current pose (reference) & -- & -- & -- & 4.90 & 1.36 & 7.79 & 0 & -- & -- \\
\midrule
ACT~\cite{zhao2023act} & 52 & 100.0k${\times}$32 & 3.3 & 3.22 & 1.95 & 4.11 & 0.750 & 93.8 & 20 \\
Diffusion Policy~\cite{chi2023diffusionpolicy} & 274 & 75.6k${\times}$64 & 6.0 & 3.21 & 2.03 & 4.07 & 0.712 & 92.7 & 960 \\
SmolVLA~\cite{shukor2025smolvla} & 450 & 27.6k${\times}$32 & 6.0 & 3.03 & 1.56 & 4.02 & 0.759 & 92.1 & 402 \\
1-loop ablation (no looping) & 199 & 76.9k${\times}$64 & 10.9 & 3.59 & 2.51 & 4.28 & 0.687 & 93.8 & 80 \\
\textbf{VeloLoop (ours, 3 loops)} & 199 & 41.8k${\times}$64 & 10.9 & 3.24 & 2.01 & 4.06 & 0.736 & 93.8 & 94 \\
\midrule
\multicolumn{10}{l}{\textit{Checkpoint selected on the periodic validation evaluation (same held-out episodes)}} \\
1-loop ablation, val.-selected & 199 & 25.0k${\times}$64 & 10.9 & 3.26 & 2.04 & 4.09 & 0.740 & 93.4 & 81 \\
VeloLoop (3 loops), val.-selected & 199 & 35.0k${\times}$64 & 10.9 & 3.22 & 1.99 & 4.04 & 0.737 & 93.8 & 94 \\
\bottomrule
\end{tabular}}
\end{table}

\paragraph{Comparison with baselines.} All learned policies reduce the chunk error well below the hold-pose reference
(\holdMAE$^\circ$, Table~\ref{tab:main}). VeloLoop (\oursMAE$^\circ$) is on par with ACT (\actMAE$^\circ$) and
Diffusion Policy (\dpMAE$^\circ$); differences of at most $0.03^\circ$ are below the run-to-run variation we would
expect, though a single seed per method does not let us measure it. SmolVLA, the only method pretrained on robot data
and the largest (450M), is the most accurate (\smolMAE$^\circ$), with its advantage concentrated in the first 8 chunk
steps (1.56$^\circ$ versus 1.95--2.03$^\circ$). At steps 33--48 the four methods lie within $0.1^\circ$. ACT is by far
the fastest (\actLat\,ms) and Diffusion Policy the slowest (\dpLat\,ms, 100 denoising steps); VeloLoop needs
\oursLatThree\,ms with 3 loops and \oursLatOne\,ms with 1. Over the first 8 steps, holding the pose (1.36$^\circ$)
beats every learned policy: chunks start with a small offset from the current state, which is harmless under
receding-horizon execution with a rate limit (Section~\ref{sec:closed-loop}).

\begin{figure}[t]
\centering
\IfFileExists{figures/training_curves.png}{\includegraphics[width=0.92\linewidth]{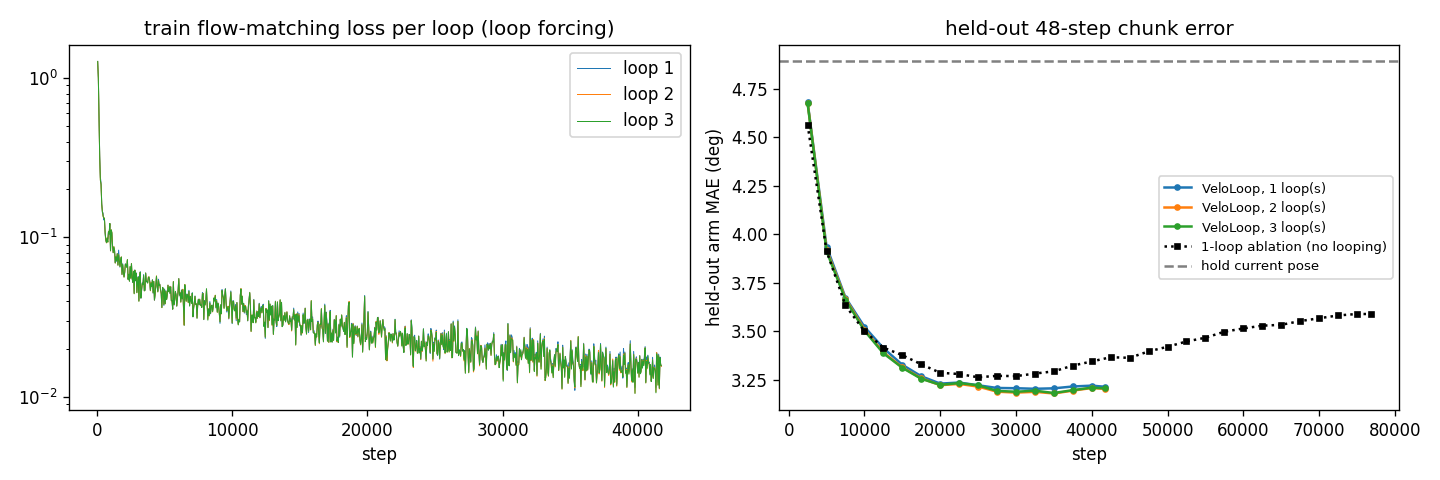}}{\fbox{training curves pending}}
\caption{Left: training flow-matching loss for each loop (the three curves overlap). Right: held-out arm MAE during
training (1{,}024 samples) for VeloLoop at each loop count and for the parameter-matched 1-loop ablation, which
overfits after about 25k steps. Per-checkpoint values are in Table~\ref{tab:curves}.}
\label{fig:curves}
\end{figure}

\paragraph{Looping as a regulariser.} The parameter-matched 1-loop ablation reaches its best validation error,
\ablPeakVal$^\circ$, at \ablBestSteps{} steps and then degrades steadily to \ablEndVal$^\circ$, while its training
loss keeps falling to roughly half of VeloLoop's (Fig.~\ref{fig:curves}). VeloLoop reaches \oursPeakVal$^\circ$ and
stays within $0.03^\circ$ of it to the end of its budget. Within the same 10.9\,h the ablation, being faster per step,
ran 1.8$\times$ more optimiser steps (\ablSteps{} versus \oursSteps{}), so we compare at equal step counts: from 20k to
40k steps VeloLoop is lower at every evaluation, by $0.04$--$0.14^\circ$ (40k: 3.21$^\circ$ versus 3.35$^\circ$;
Table~\ref{tab:curves}), at a similar training loss. On the full evaluation the final models score \oursMAE$^\circ$
versus \ablMAE$^\circ$, and even validation-selected checkpoints, which favour the overfitting ablation, give
\oursBestMAE$^\circ$ versus \ablBestMAE$^\circ$. We read this as loop forcing regularising the shared stack: it must be
useful after every pass, and every sample receives three supervised readouts. Two confounds remain: with wall-clock
schedules the ablation's learning rate was higher at any given step, and both runs use a single seed.

\begin{table}[t]
\centering
\caption{One VeloLoop checkpoint evaluated with 1, 2 or 3 loops. Latency: median batch-1 time including the 10 flow
steps, on a shared GPU.}
\label{tab:loops}
\begin{tabular}{crrrrr}
\toprule
Loops at inference & MAE$\downarrow$ & MAE 1--8$\downarrow$ & Grip.\ acc.\ (\%) & Encoder (ms) & Total (ms) \\
\midrule
1 & 3.24 & 2.01 & 93.8 & 14 & 84 \\
2 & 3.24 & 2.02 & 93.9 & 22 & 87 \\
3 & 3.24 & 2.01 & 93.8 & 30 & 94 \\
\bottomrule
\end{tabular}
\end{table}

\paragraph{Loops at inference.} Evaluated with 1, 2 or 3 loops, the same checkpoint gives the same accuracy
(\oursOneLoopMAE$^\circ$ at every loop count, Table~\ref{tab:loops}): the second and third loops add nothing on this
dataset, and the three loops' training losses are indistinguishable (Fig.~\ref{fig:curves}). Each loop costs about
8\,ms of encoder time (\oursEncOne\,ms for one, \oursEncThree\,ms for three). Looping is therefore useful here as
training-time regularisation, not as test-time compute, and loop forcing is what makes it free at inference: the model
can be deployed with one loop at the cost of a plain 6-layer encoder. Whether extra loops help on harder or more
diverse tasks remains open. A replay of held-out episodes through the exact deployment preprocessing (raw videos,
live-camera resizing, the released checkpoint) gives an arm MAE of \replayMAE$^\circ$ against \replayHold$^\circ$ for
holding the pose, with one clear failure in which the policy reaches for a different block position
(Appendix~\ref{app:replay}).

\subsection{Closed-loop evaluation on the robot}
\label{sec:closed-loop}
We ran the released checkpoint (\oursBestSteps{} steps, 10 flow steps) on the 6-DoF manipulator with ``pick the yellow block
put into the box''. It was deployed in its released 3-loop configuration; since one loop matches three offline
(Table~\ref{tab:loops}), a single-loop deployment would save about 16\,ms of encoder time per chunk, which was not
needed here. The 30\,Hz control loop reads the joint state over CAN and both cameras directly, and executes each
48-step chunk receding-horizon: a new chunk is requested when 16 steps remain, and its first steps are skipped by the
ticks executed while it was computed. Commands pass a per-tick rate limit and the joint limits, the gripper is
binarised, and the servos run at 50\,\% speed. Plan playback pauses if the arm lags by more than 0.35\,rad, and the arm
holds whenever a frame is older than 200\,ms. Inference runs on a GPU on the robot computer.

We report one continuous 11.0-minute session (19{,}740 ticks) without restarting the policy; the block was returned
to the table after each placement. Grasps are labelled automatically from gripper telemetry. An attempt is a period
with the gripper commanded closed while the arm is on the pick side; it \emph{held} the block if the jaws stopped at
the block's width (50--55\,mm), \emph{missed} if they closed fully ($<$15\,mm), and \emph{slipped} if they closed
below 15\,mm while still commanded shut. Every release happened with the base joint at $+0.43$ to $+0.77$\,rad (the box
side; grasps began at $-0.50$ to $+0.02$\,rad), so we count releases as placements at the box. As in the
demonstrations, the policy closes the gripper again after each release; these closings are not grasps. Without a
camera label we cannot tell whether a block rested inside the box or on its rim.

\begin{table}[t]
\centering
\caption{Closed-loop robot session: one uninterrupted 11.0-minute run at 30\,Hz with the released checkpoint.
Outcomes are labelled from gripper telemetry (see text). Latency is the policy's time per 48-step chunk on the robot
computer; frame age is the age of the oldest input when a chunk was requested; holding ticks have no plan, a stale
frame or a lagging arm.}
\label{tab:closed-loop}
\small
\begin{tabular}{lr@{\hspace{2.5em}}lr}
\toprule
Grasp attempts & 25 & Time between placements (median / mean) & 28\,s / 44\,s \\
\quad held the block & 17 (68\,\%) & Chunk latency (median / p95) & 35\,ms / 44\,ms \\
\quad\quad placed at the box & 14 (82\,\% of held) & Input frame age (median) & 23\,ms \\
\quad\quad slipped & 3 (18\,\% of held) & Ticks holding pose & 15 of 19{,}740 \\
\quad missed & 8 (32\,\%) & & \\
\bottomrule
\end{tabular}
\end{table}

\begin{figure}[t]
\centering
\includegraphics[width=\linewidth]{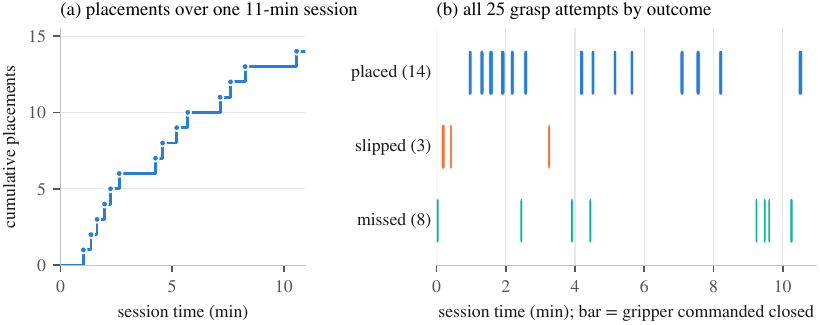}
\caption{The closed-loop session. (a) Cumulative placements at the box. (b) Every grasp attempt, from gripper
close to open command, by outcome.}
\label{fig:closed-loop}
\end{figure}

The policy kept working for the whole run (Table~\ref{tab:closed-loop}, Fig.~\ref{fig:closed-loop}): placements are
spread across the session, and after a miss or slip it re-approached the block without intervention. Of 25 grasp
attempts, 17 held the block and 14 of those reached the box, one placement every 28\,s (median). The main failure is
the grasp itself: 8 attempts closed on nothing, consistent with the mislocalised reach seen in the replay. The
control loop was not a bottleneck: 35\,ms per chunk, 23\,ms-old frames and holding on 0.1\,\% of ticks. This is one
session with one instruction and no closed-loop baseline, so we read it as evidence that the offline accuracy
translates into a working closed-loop policy, not as a success rate to rank methods.

\section{Limitations}
The main evaluation is offline and open-loop. The closed-loop result is a single session with one instruction,
labelled from gripper telemetry rather than by a camera or a human, and no baseline has been run closed-loop yet; it
also used the 3-loop configuration rather than the single loop we recommend. All results come from one robot, one
dataset with three closely related instructions and a single training seed. Training shared a GPU with other jobs, so
budgets and step counts differed between methods; the baselines use untuned LeRobot defaults, and only SmolVLA is
pretrained on robot data. The looped-versus-non-looped comparison controls for step count but not for the different
learning rate at a given step under wall-clock schedules. Extra loops did not improve accuracy at inference, so the
benefit of looping shown here is regularisation; test-time compute scaling with loops remains to be shown. Loop
memory is detached from the gradient, as in the reference implementation.

\FloatBarrier
\section{Conclusion}
VeloLoop brings mixture-of-loops attention to VLA policies by looping a weight-shared encoder over the observation
context and using a cross-only MoLT decoder as a flow-matching action expert. On 1{,}116 real-robot demonstrations it
matches ACT and Diffusion Policy in held-out chunk accuracy and trails the robot-pretrained SmolVLA. Its clearest
benefit is regularisation: looped training with loop forcing removes the overfitting of a parameter-matched
non-looped model, and because every loop is supervised, this comes at the inference cost of a single loop. On the
real arm it ran an uninterrupted 11-minute closed-loop session at 35\,ms per chunk and placed the block 14 times in
25 grasp attempts, with grasp localisation as the main failure. Closed-loop baselines, single-loop deployment on the
robot, multiple seeds and harder tasks, where extra loops may pay off, are the next steps.

\appendix
\section{Implementation and training details}
\label{app:impl}
\paragraph{Fused MoLA.} Depth sources differ for every query, so MoLA cannot be written as standard attention. We
concatenate the sequence keys and the $\ell+1$ depth slots into one key set of length $N(\ell+2)$ and use a boolean
mask that lets query $t$ see sequence keys $j\neq t$ and only entry $t$ of each depth slot. This computes exactly the
joint softmax of Eq.~(1) in a single fused attention call (maximum deviation $2\times10^{-7}$ from an explicit
implementation) and was about $2\times$ faster.

\paragraph{Training.} VeloLoop has 198.8M parameters, 155.6M trainable. We use AdamW ($\beta=(0.9,0.95)$, weight decay
0.05, learning rate $2\times10^{-4}$, $0.1\times$ for the vision tower), 1k warm-up steps, cosine decay, batch 64, bf16,
gradient clipping at 1.0 and an EMA of the weights (0.999), used for evaluation. Augmentation: random shift up to 12
pixels, brightness, contrast and saturation jitter; state noise 0.05 and state dropout 0.1. Training ran on a single
cloud NVIDIA RTX A6000 GPU (48\,GB), shared with other jobs, at 0.55--1.0\,s per step depending on contention, using
about 22\,GB with gradient checkpointing. ACT ran for \actSteps{} steps (batch 32), Diffusion Policy for \dpSteps{} (batch 64), SmolVLA for
\smolSteps{} (batch 32), VeloLoop for \oursSteps{} and the 1-loop ablation for \ablSteps{} (both batch 64).

\begin{table}[h]
\centering
\caption{Looped versus non-looped training. Validation MAE every 2{,}500 steps on 1{,}024 fixed held-out samples (3
loops for VeloLoop); train FM loss is the loop-1 flow-matching loss averaged over $\pm$2k steps. Both runs had the
same 10.9\,h budget; the 1-loop ablation is faster per step and so ran longer.}
\label{tab:curves}
\begin{tabular}{rrrrr}
\toprule
Step & \multicolumn{2}{c}{Val.\ MAE (deg)$\downarrow$} & \multicolumn{2}{c}{Train FM loss (loop 1)} \\
 & VeloLoop (3 loops) & 1-loop ablation & VeloLoop & 1-loop \\
\midrule
10k & 3.50 & 3.50 & 0.0359 & 0.0367 \\
20k & 3.22 & 3.29 & 0.0239 & 0.0260 \\
25k & 3.22 & 3.26 & 0.0236 & 0.0252 \\
30k & 3.19 & 3.27 & 0.0193 & 0.0193 \\
35k & 3.18 & 3.29 & 0.0190 & 0.0168 \\
40k & 3.21 & 3.35 & 0.0148 & 0.0173 \\
\midrule
end of budget (41.8k / 76.9k) & 3.21 & 3.59 & 0.0145 & 0.0073 \\
\bottomrule
\end{tabular}
\end{table}

\paragraph{Per instruction.} VeloLoop is most accurate on ``pick the yellow block'' (\instrBMAE$^\circ$, \instrBN{}
samples), a short task without the placing phase, and least accurate on ``pick the black block put into the box''
(\instrAMAE$^\circ$, \instrAN{} samples), which has 159 training episodes against 854 for the yellow-block variant
(\instrCMAE$^\circ$, \instrCN{} samples).

\section{Ground-truth replay through the deployment pipeline}
\label{app:replay}
\begin{figure}[!h]
\centering
\IfFileExists{figures/replay_example.png}{\includegraphics[width=0.43\linewidth]{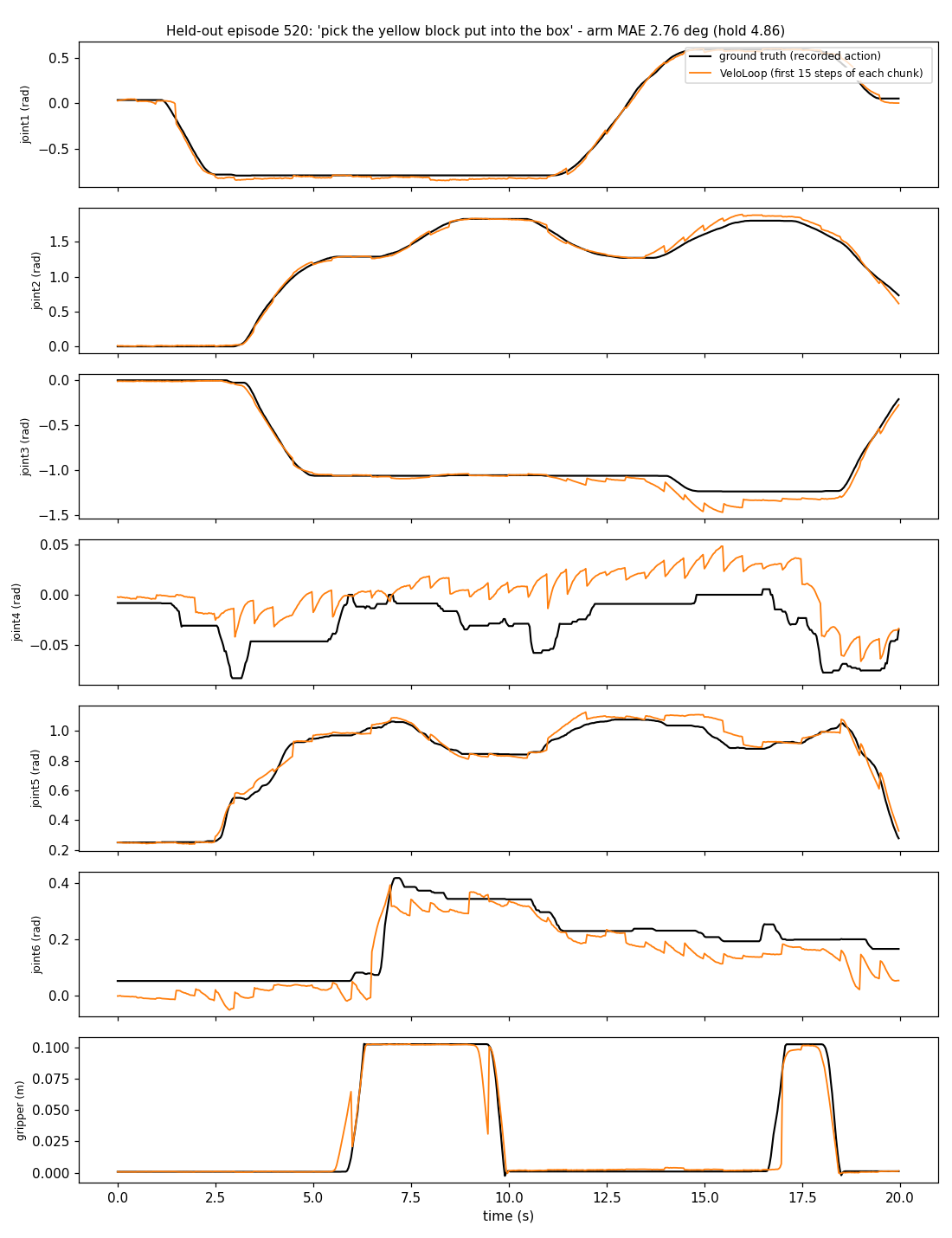}}{\fbox{replay figure pending}}\hfill
\IfFileExists{figures/replay_failure.png}{\includegraphics[width=0.43\linewidth]{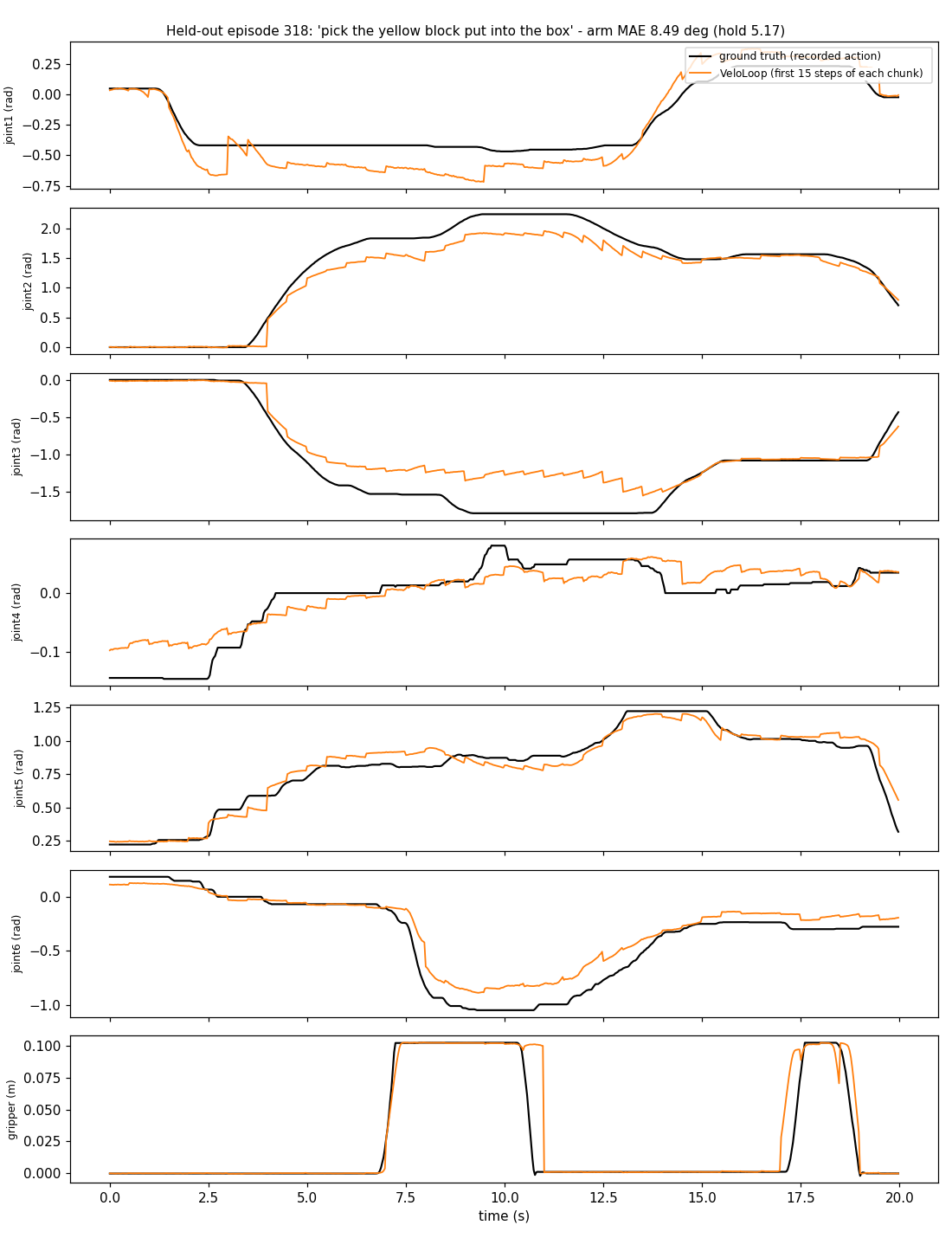}}{}
\caption{Held-out episodes replayed from the raw videos. Left: a typical episode (\replayTypEp, median error,
\replayTypMAE$^\circ$). Right: the worst episode (\replayWorstEp, \replayWorstMAE$^\circ$), where the policy aims at a
different block position during the reach. Black: recorded action; orange: the first 15 steps of each VeloLoop chunk
(re-planned every 0.5\,s). Joints in radians; gripper value is the jaw opening in metres.}
\label{fig:replay}
\end{figure}

The cached-frame evaluation shares its preprocessing with training. As a final check before the robot, we replay
held-out episodes through the exact deployment path: raw frames are decoded from the original $1280\times720$ and
$640\times480$ videos and resized as the live camera images are, each episode uses its recorded instruction, and the
released stand-alone script runs the released, validation-selected checkpoint (\oursBestSteps{} steps). A chunk is
predicted every 15 frames (0.5\,s) and compared with the recorded actions. Over \replayN{} episodes the arm MAE is
\replayMAE$^\circ$, against \replayHold$^\circ$ for holding the pose (Table~\ref{tab:replay}). Five episodes track the
demonstration closely (2.4--3.7$^\circ$) and the gripper state agrees with the recording on 88--95\% of frames.
Episode 318 is a clear failure: throughout the reach and grasp the predicted base-joint target is offset by about
0.2\,rad, i.e.\ the policy aims at a different block position, and its error (8.49$^\circ$) exceeds the hold reference
(Fig.~\ref{fig:replay}). This mislocalisation is what an offline average hides, and it matches the missed grasps on
the robot.

\begin{table}[h]
\centering
\caption{Ground-truth replay of held-out episodes from raw videos through the deployment preprocessing. MAE is over
the six arm joints in degrees for full 48-step chunks predicted every 0.5\,s.}
\label{tab:replay}
\resizebox{0.8\linewidth}{!}{\begin{tabular}{rlrrr}
\toprule
Episode & Instruction & Frames & VeloLoop MAE$\downarrow$ & Hold MAE \\
\midrule
912 & pick the yellow block & 300 & 2.40 & 3.31 \\
707 & pick the yellow block put into the box & 600 & 2.41 & 3.81 \\
852 & pick the yellow block put into the box & 600 & 2.80 & 4.00 \\
520 & pick the yellow block put into the box & 600 & 2.76 & 4.86 \\
318 & pick the yellow block put into the box & 600 & 8.49 & 5.17 \\
70 & pick the yellow block put into the box & 600 & 3.72 & 7.52 \\
\midrule
Mean & & & 3.76 & 4.78 \\
\bottomrule
\end{tabular}}
\end{table}

\FloatBarrier
% bibliography inlined (plainnat, from refs.bib)

\end{document}